\documentclass[pdflatex,sn-mathphys-num]{sn-jnl}
\usepackage[T1]{fontenc}
\usepackage{hyperref}
\usepackage{url}
\usepackage{amssymb,amsfonts}
\usepackage[all,arc, cmtip]{xy}
\usepackage{enumerate}
\usepackage{caption}
\usepackage{subcaption}
\usepackage{tikz}
\usetikzlibrary{arrows,decorations.pathmorphing,backgrounds,positioning,fit,petri}
\usetikzlibrary{decorations.pathreplacing,calligraphy}
\usepackage{paralist}
\usepackage{booktabs} 
\usepackage{array} 
\usepackage{mathrsfs}
\usepackage{url}
\usepackage{color}
\usepackage{cancel}
\usepackage{todonotes} 
\usepackage{hyperref} 
\usepackage{mathtools}
\usepackage{multirow}
\usepackage[mathscr]{euscript}
\usepackage[nameinlink,capitalise,noabbrev]{cleveref}
\usepackage[shortcuts]{extdash}
\usepackage[normalem]{ulem}
\usepackage{enumitem}

\newcommand{\Feat}{\mathrm{Feat}}

\usepackage{tikz-cd}

\newtheorem{thm}{Theorem}[section]
\newtheorem*{thm*}{Theorem}

\newtheorem{prop}[thm]{Proposition}

\theoremstyle{definition}

\newtheorem{defn}[thm]{Definition}
\newtheorem*{defn*}{Definition}

\newtheorem{rem}[thm]{Remark}

\theoremstyle{remark}

\newcommand{\TS}{\mathrm{TS}}
\newcommand{\RR}{\mathbb{R}}
\newcommand{\Z}{\mathbb{Z}}

\title{GNN-Enhanced TCN Algorithms for ECG Anomaly Detection}

\author[1]{\fnm{Angelica} \sur{Simonetti}}\email{angelica.simonetti@gmail.com}\equalcont{All authors contributed equally to this work.}

\author*[1]{\fnm{Ferdinando} \sur{Zanchetta}}\email{ferdinando.zanchett2@unibo.it}
\equalcont{All authors contributed equally to this work.}

\tikzstyle{nd1}=[circle,draw=blue!50,fill=blue!10!,thick,
inner sep=0pt,minimum size=3mm, rounded corners=1ex]
\tikzstyle{nd2}=[circle,draw=blue!50,fill=blue!10!,thick,
inner sep=0pt,minimum size=3mm, rounded corners=1ex]
\tikzstyle{nd3}=[circle,draw=blue!50,fill=blue!10!,thick,
inner sep=0pt,minimum size=3mm, rounded corners=1ex]

\tikzstyle{inout}=[rectangle,draw=black!50,fill=black!10,thick,
inner sep=0pt, minimum width=12mm, minimum height=5mm, rounded corners=1ex]
\tikzstyle{layer}=[rectangle,draw=blue!50,fill=blue!10,thick,
inner sep=0pt, minimum width=20mm, minimum height=5mm, rounded corners=1ex]
\tikzstyle{main}=[rectangle,draw=black!50!green,fill=black!20!green!15,thick,
inner sep=0pt, minimum width=20mm, minimum height=5mm, rounded corners=1ex]
\tikzstyle{bn}=[circle,draw=red!50,fill=red!10,thick,
inner sep=0pt,minimum size=6mm, rounded corners=1ex]
\tikzstyle{bg}=[fill=black!5!yellow!20, minimum width=25mm, minimum height=20mm, rounded corners=1ex]
\tikzstyle{bg2}=[fill=black!5!green!20, minimum width=25mm, minimum height=20mm, rounded corners=1ex]
\tikzstyle{bg3}=[fill=black!5!yellow!20, minimum width=25mm, minimum height=29mm, rounded corners=1ex]
\tikzstyle{times}=[rectangle,draw=black!80!yellow!60, fill=black!20!yellow!20, minimum width=5mm, minimum height=5mm, rounded corners=1ex]
\tikzstyle{times2}=[rectangle,draw=black!80!green!60, fill=black!20!green!20, minimum width=5mm, minimum height=5mm, rounded corners=1ex]

\begin{document}
\abstract{Temporal convolutional networks (TCNs) are among the most effective algorithms to deal with time series data.
To a time series can be also given the structure of directed graph, opening the doors to the usage of graph neural networks (GNNs) in this context. In this paper we develop a new pipeline involving novel geometric deep learning models that merge the capabilities of ordinary TCNs with the ones of GNNs to solve an anomaly detection problem on electrocardiogram signals.}
\maketitle

\section{Introduction}
Over the last few years, temporal convolutional networks (TCNs) gained prominence as one of the most effective classes of deep learning (DL) algorithms (see \cite{LeCun2015}, \cite{Bai2018}) to deal with time series data (\cite{HamiltonTS}), in many cases surpassing the effectiveness of more traditional DL algorithms like recurrent neural networks (RNNs, comprising Long Short-Term Memory networks aka LSTMs).
On the other hand, when dealing with graph structured data, a particular class of geometric deep learning (see \cite{Bronstein2016GeometricDL}) algorithms called graph neural networks (GNNs, see \cite{Bronstein2016GeometricDL}, \cite{GNNSurvey19}) has proven to be very effective at solving problems related with this type of data or at improving the capabilities of existing algorithms. In the context of time series, GNNs have been used jointly with time series specific architectures (e.g. RNNs or TCNs), for example, to study auto-regressive models arising from graph time series or to forecast traffic speed or density (see the survey \cite{GNNSurvey19}). In the latter case, the graph structure involved is given by the positions of the sensors that record time series, see \cite{Graphwavenet19}, or it is inferred by learning the weights of an adjacency matrix (\cite{Graphwavenet19}).
\par
In this paper we explore a different and novel approach, that is to use the graph structure to encode time dependencies. We presented some preliminary findings based on this approach on the occasion of the conference GSI2025 (\cite{GSI2025}). Our idea is to introduce a \emph{directed} graph structure where the information associated to a given time step (a node) can be linked with the information coming from time steps in the past (this relation assumes the form of a directed edge). This allows us to couple the advantages of TCNs with the ones of graph convolutions. We prove that this approach enhances state-of-the-art TCNs architectures and we test it on the problem of identifying anomalies in electrocardiogram (ECG) signals. We shall now summarize the contents of the paper. 
\par
In the first part of the paper (Sections \ref{sec:TCN} and \ref{sec:gnn}) we introduce temporal convolutions and graph convolutions from a mathematical viewpoint and we study how to see a time series as a directed graph. We explain how to see temporal convolutions as graph convolutions on directed digraphs thus proving that TCNs can be seen as GNNs for a certain choice of digraph structure (Proposition \ref{1dconv}). As a consequence, we motivate in these sections why graph convolutions might be useful for the study of TCNs: by definition, as we shall explain, they exhibit a so called \emph{permutation equivariance} that seems to be particularly relevant when dealing with time dependencies as the ones arising into the context of time series.
\par
Motivated by our theoretical observations, we describe in Section \ref{sec:models} some DL models that we designed and used to deal with time series data. These are autoencoders employing both temporal convolutions (that can be seen as graph convolutions) and genuine graph convolutions.
\par
In Section \ref{sec:exp} we describe the experiments we carried out on real world datasets to test the effectiveness of our algorithms. In this paper, we considered problems dealing with ECG signals, that are signals recording the electrical activity of the heart. They can be used to monitor the health status of a human being and to diagnose various heart conditions. In addition, they are easier to acquire than other medical data as they can be recorded by non-invasive devices. Recordings of ECG signals may present many types of anomalies, going from biological ones (e.g. arrhythmias) to artifacts linked with recording issues (e.g corruption due to device malfunctioning, environmental noise, incorrect usage of the device, etc.). We have applied our algorithms to solve an anomaly detection problem on an established benchmark dataset, going beyond the quality detection problem we solved in \cite{GSI2025}: the ECG5000 dataset. This dataset only contains genuine signals that can be normal or abnormal: we establish that GNN-enhanced autoencoders achieve state-of-art performance and allow the design of lightweight models (reducing the number of learnable parameters) that can possibly be deployed more easily than the ones already developed in the literature we consulted. This approach is different from other classical methods relying on hand crafted features in what it considers only raw signals as an input. We also study the performance of variational autoencoders in this context. 
\par
Appendix \ref{hyp} contains the technical details about our software implementations for reproducibility and a summary of the main quantitative results obtained in \cite{GSI2025} to complement our discussion.
\par
\textbf{Acknowledgements:} Both authors wish to thank Andrea Malagoli, Andrea Arigliano and Giovanni Faglioni for sharing with us their expertise about ECG signals. We also thank Rita Fioresi and Pietro Demurtas for stimulating discussions.

\section{Temporal Convolutional Networks}\label{sec:TCN}
Convolutional neural networks (CNNs, see \cite{lbbh}, \cite{lecun2010mnist}, \cite{kr-imagenet}, \cite{LeCun2015}) are a type of deep learning architecture that has been introduced to process grid-like data, such as 2D images. It relies on specific layers called \emph{convolutional layers} which apply learnable sliding filters to local regions of the inputs. CNNs have been also adapted to data organized in sequences (1D CNNs, see \cite{1DCNN19}) and, even more specifically, to time series (TCNs, see \cite{Bai2018} and the references therein). Let us start defining time series and establishing the notation we will use throughout the paper.

\begin{defn}
    We denote by $\TS(\Z,m)$ the set of functions $\Z\rightarrow \RR ^m$. Let $\TS(r,m)$ be the set of multivariate time series with $m$ channels and of length $r$, that is the set of functions from the discrete ordered set $[r]:=[0,...,r-1]\subseteq\Z$ to $\mathbb{R}^m$. Given $\textbf{x}\in\TS(r,m)$ we will denote as $\textbf{x}(i)_j$ (or simply as $\textbf{x}_{ij}$ when no confusion is possible), for $i=0,...,r-1$ and $j=1,...,m$, the $j$th coordinate of the vector $\textbf{x}(i)\in\RR^m$.
\end{defn}
As we anticipated, TCNs are built on 1D convolutions.
\begin{defn}\label{def:conv}
    Given a filter $K\in\RR^f$, we can define a one-channel, one-dimensional (1D) convolution on $\TS(\Z,1)$ as an operator
    \begin{align*}
    &\mathrm{conv1D}:\TS(\Z,1)\rightarrow\TS(\Z,1) \\
    &\mathrm{conv1D}(\textbf{x})(j)=\sum_{i=0}^{f-1}K_i\textbf{x}(\alpha(j,i))+b
\end{align*}
where $\alpha(j,-):[f]\rightarrow\mathbb{Z}$ are strictly monotonic index functions. Similarly, one obtains operators
\begin{align*}
    &\mathrm{conv1D}:\TS(r,1)\rightarrow\TS(l,1) \\
    &\mathrm{conv1D}(\textbf{x})(j)=\sum_{i=0}^{f-1}K_i\textbf{x}(\alpha(j,i))+b
\end{align*}
where $\alpha(j,-):[f]\rightarrow\mathbb{Z}$ are strictly monotonic index functions, $\textbf{x}(i):=0$ if $i\notin [r]$ and $b\in\mathbb{R}$ is called bias.
The numbers $K_i$, $b$ are called the \emph{parameters} or \emph{weights} of the convolution.
\end{defn}
The most commonly used index functions are of the form $\alpha(j,i)=(n+d\cdot i)+j$ for some integers $n,d$. As a consequence, from now on we shall assume that the one dimensional convolutions we consider have this form, unless otherwise stated. If, $\alpha(j,i)\leq j$ for all $i,j$ then the convolution is said to be \emph{causal} as it will look only 'backward'. If $|d|>1$, the convolution is called \emph{dilated}. Dilated causal convolutions are the building blocks of TCNs, that are CNNs where only causal convolutions appear, as we shall describe below.
We define multi-channel (i.e. handling multivariate time series) 1D convolutions, in two steps
\begin{defn}
    We define operators $$\mathrm{conv1D}:\TS(r,n)\rightarrow\TS(l,1),\qquad\mathrm{conv1D}(\textbf{x})(i)=\sum_{j=1}^n\mathrm{conv1D}_j(\textbf{x}(-)_{j})(i)$$
    where $\mathrm{conv1D}_j: \TS(r,1)\rightarrow\TS(l,1)$ are $n$ different convolutions, and operators 
    $$\mathrm{conv1D}:=\oplus\mathrm{conv1D}:\TS(r,n)\rightarrow\TS(l,m)\cong\oplus\TS(l,1)$$
\end{defn}
\begin{rem}
    Similar considerations apply to define convolutional operators $\TS(\Z,n)\rightarrow\TS(\Z,m)$
\end{rem}
We can now define temporal convolutions as follows.
\begin{defn}
    We define temporal convolutions to be dilated 1D convolutions $\TS(r,n)\rightarrow\TS(r,m)$ (or more generally $\TS(\Z,n)\rightarrow\TS(\Z,m)$), that is, dilated 1D convolutions transforming a multivariate discrete time series of a given length into a multivariate discrete time series of the same length. Convolutional neural networks are neural networks that employ a sequence of convolutional and and pooling layers alternated with activation functions, that is multivariate ridge functions. TCNs are defined as the subclass of CNNs that take as an input series $\textbf{x}\in\TS(r,n)$ and that employ as convolutional layers only temporal convolutions.
\end{defn}
The usefulness of TCNs in the context of time series arises from the fact that causal convolutions by design are able to exploit temporal dependencies, while not suffering from some of the algorithmic problems of RNNs such as LSTMs: for example they appear to be faster to train and more scalable (see \cite{Bai2018} for a discussion). It is often stated that \emph{CNNs are translation invariant}. In our context, we can give to this statement a more precise mathematical meaning.
\begin{prop}
    The translation group $T_\Z\cong(\Z,+)$ acts on $\TS(\Z,n)$ as follows:
    $$T_a(\textbf{x})=\textbf{x}\circ T_a\qquad \textbf{x}\in\TS(\Z,n),\quad a\in\Z$$
    where $T_\Z\ni T_a:\Z\rightarrow\Z$, $T_a(y)=y+a$. Consider a convolution $\mathrm{conv1D}:\TS(\Z,1)\rightarrow\TS(\Z,1)$ as in \ref{def:conv}. Then if for all $j\in\Z$ and for all $T_a\in T_\Z$ we have $T_a(\alpha(j,-))=\alpha(T_a(j),-)$ we have that $\mathrm{conv1D}$ is $T_\Z$ equivariant. The same applies to the case of convolutions $\TS(\Z,n)\rightarrow\TS(\Z,m)$.
\end{prop}
\begin{proof}
    We only notice that, under the hypothesis of the statement, 
    \begin{align*}T_a(\mathrm{conv1D}(\textbf{x}))(j)&=\mathrm{conv1D}(\textbf{x})(T_a(j))=\sum_{i=0}^{f-1}K_i\textbf{x}(\alpha(T_a(j),i))+b= \\
    &=\sum_{i=1}^fK_i\textbf{x}(T_a(\alpha(j,i)))+b =\\
    &=\sum_{i=1}^fK_iT_a(\textbf{x})(\alpha(j,i))+b=\mathrm{conv1D}(T_a(\textbf{x}))(j)
    \end{align*}
\end{proof}

\begin{rem}\label{rem:tcnlimits}
    Explicitly note that:
    \begin{itemize}
        \item Convolutions having $\alpha(j,i)=(n+d\cdot i)+j$  for some $n,d\in\Z$ satisfy the hypothesis of the previous proposition.
        \item The action of $T_\Z$ does not restrict trivially to an action on $\TS(l,n)$, for a given natural number $l$. As a consequence, in this case, the mathematical notion of \emph{translation invariance} for 1D (temporal) convolution of the form $\TS(l,n)\rightarrow\TS(l,m)$, for example, must be adapted.
    \end{itemize}
\end{rem}

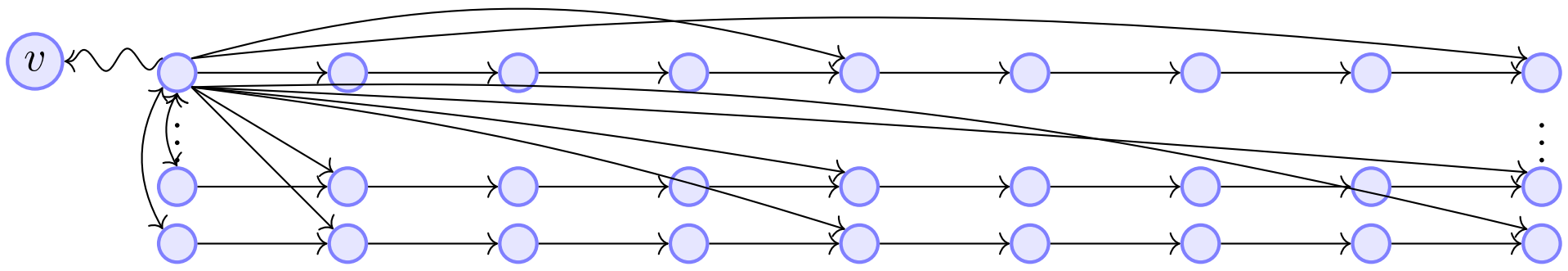
\begin{figure*}[t]
\vspace*{-0.5cm}
\begin{center}
\begin{tikzpicture}[scale=0.46]
\node at (0,0) [nd1] (p11) {};
\node at (1*3,0) [nd1] (p12) {};
\node at (2*3,0) [nd1] (p13) {};
\node at (3*3,0) [nd1] (p14) {};
\node at (4*3,0) [nd1] (p15) {};
\node at (5*3,0) [nd1] (p16) {};
\node at (6*3,0) [nd1] (p17) {};
\node at (7*3,0) [nd1] (p18) {};
\node at (8*3,0) [nd1] (p19) {};

\draw [->] (p11.east) -- (p12.west);
\draw [->] (p12.east) -- (p13.west);
\draw [->] (p13.east) -- (p14.west);
\draw [->] (p14.east) -- (p15.west);
\draw [->] (p15.east) -- (p16.west);
\draw [->] (p16.east) -- (p17.west);
\draw [->] (p17.east) -- (p18.west);
\draw [->] (p18.east) -- (p19.west);

\node at (0,1) [nd2] (p21) {};
\node at (1*3,1) [nd2] (p22) {};
\node at (2*3,1) [nd2] (p23) {};
\node at (3*3,1) [nd2] (p24) {};
\node at (4*3,1) [nd2] (p25) {};
\node at (5*3,1) [nd2] (p26) {};
\node at (6*3,1) [nd2] (p27) {};
\node at (7*3,1) [nd2] (p28) {};
\node at (8*3,1) [nd2] (p29) {};

\draw [->] (p21.east) -- (p22.west);
\draw [->] (p22.east) -- (p23.west);
\draw [->] (p23.east) -- (p24.west);
\draw [->] (p24.east) -- (p25.west);
\draw [->] (p25.east) -- (p26.west);
\draw [->] (p26.east) -- (p27.west);
\draw [->] (p27.east) -- (p28.west);
\draw [->] (p28.east) -- (p29.west);

\node at (8*3,2) {$\vdots$};
\node at (0,2) {$\vdots$};

\node at (-1*2.5,3.2) [nd3] (e0) {$\ v\ $};
\node at (0,3) [nd3] (e1) {};
\node at (1*3,3) [nd3] (e2) {};
\node at (2*3,3) [nd3] (e3) {};
\node at (3*3,3) [nd3] (e4) {};
\node at (4*3,3) [nd3] (e5) {};
\node at (5*3,3) [nd3] (e6) {};
\node at (6*3,3) [nd3] (e7) {};
\node at (7*3,3) [nd3] (e8) {};
\node at (8*3,3) [nd3] (e9) {};

\draw [->] (e1.east) -- (e2.west);
\draw [->] (e2.east) -- (e3.west);
\draw [->] (e3.east) -- (e4.west);
\draw [->] (e4.east) -- (e5.west);
\draw [->] (e5.east) -- (e6.west);
\draw [->] (e6.east) -- (e7.west);
\draw [->] (e7.east) -- (e8.west);
\draw [->] (e8.east) -- (e9.west);

\draw[->, decorate,decoration=snake] (e1.north west) -- (e0.east);

\path [->] (e1.north east) edge [bend left=15]  (e5.north west);
\path [->] (e1.north east) edge [bend left=6]  (e9.north west);
\draw [<->] (e1.south) edge [bend right=30] (p21.north);
\path [<->] (e1.south west) edge [bend right=30] (p11.north west);
\draw [->] (e1.south east) -- (p22.north west);
\draw [->] (e1.south east) -- (p12.north west);
\path [->] (e1.south east) edge [bend left=2]  (p25.north west);
\path [->] (e1.south east) edge [bend left=1]  (p29.north west);
\path [->] (e1.south east) edge [bend left=5]  (p15.north west);
\path [->] (e1.south east) edge [bend left=7.5]  (p19.north west);
\end{tikzpicture}
\end{center}
\caption{\footnotesize{One possible structure of a time-digraph, as described in Example 1. Here only adjacent connections and all the connections for node $v$ are shown, and $d=4$}}\label{tgraph}
\end{figure*}
\section{Time series as Directed Graphs}\label{sec:gnn}
\subsection{Generalities.}
\begin{defn}
    A \emph{directed graph (digraph)} $G$ is the datum $G=(V_G,E_G,h_G,t_G)$ of two sets $V_G$ (the \emph{set of vertices}), $E_G$ (the \emph{set of edges}) and two functions $h_G,t_G:E_G\rightarrow V_G$ associating to each edge $e$ its \emph{head} $h_G(e)$ and its \emph{tail} $t_G(e)$ respectively. A morphism $\varphi:G\rightarrow H$ between two directed digraphs $G$ and $H$ is the datum of two functions $\varphi_V:V_G\rightarrow V_H$, $\varphi_E:E_G\rightarrow E_H$ such that $h_H\circ\varphi_E=\varphi_V\circ h_G$ and $t_H\circ\varphi_E=\varphi_V\circ t_G$
\end{defn}
From now on, for simplicity we will assume that our digraphs have at most one edge connecting two different nodes (for each direction) and at most one self loop for each node. In this case, given a digraph $G=(V_G,E_G,h_G,t_G)$ and an ordering of the vertices $(v_i)_{i\in [|V_G|]}$, we can define the \emph{adjacency matrix of $G$} as the matrix defined by $A_{ij} = 1$ if there is there exists an edge having as tail $v_i$ and head $v_j$ or $A_{ij} = 0$ otherwise. If the adjacency matrix of a graph is symmetric, we say that our graph is \emph{undirected}. We can assign \emph{weights} to the edges of a graph by letting the entries of the adjacency matrix to be arbitrary real numbers. When speaking about weighted digraphs, we shall always assume that there is an edge having as tail $v_i$ and head $v_j$ if and only if $A_{ij} \neq 0$. We speak in this case of \emph{weighted digraphs}. 
\begin{defn}
    A \emph{digraph with features of dimension $n$} is the datum $(G,F_G=(h_{v})_{v\in V_G})$ of a (weighted or unweighted) digraph $G$ and vectors $h_v\in\RR^n$ of \emph{node features} for each vertex $v\in V_G$. For a given digraph $G$, we shall denote as $\Feat(G,n)$ the set of all digraphs with features of dimension $n$ having $G$ as underlying digraph.
\end{defn}
Real-world graph-structured datasets usually come in the form of one or more digraphs with features. Graph Neural Networks (GNNs, \cite{GNNSurvey19}) are models that are employed on graph-structured data using as building blocks the so called \emph{graph convolutions} (\cite{Bronstein2016GeometricDL}, \cite{Fioresi2023DeepLA}): given a graph, they update each node feature vector combining the information contained in the feature vectors of adjacent nodes. In general, a graph convolution is a function $\mathrm{gconv}:\Feat(G,n)\rightarrow\Feat(G,m)$. Recall that the symmetric group $S_n$ of permutations of $n$ elements acts by conjugacy on the set of real valued $n\times n$ matrices $\mathrm{M}_{n,n}(\RR)$. Indeed, if for a given $\pi\in S_n$ we denote with $P_\pi$ the matrix permutation matrix having as entries $(P_\pi)_{ij}=\delta_{i,\pi(j)}$ we have that, if $A\in\mathrm{M}_{n,n}(\RR)$, $P_\pi A$ and $AP_\pi^T$ correspond to the matrix $A$ where the rows or the columns respectively of $A$ have been swapped according to $\pi$. Remembering that $P_\pi^{-1}=P_\pi^T$ for all $\pi\in S_n$ acts then by conjugation on $\mathrm{M}_{n,n}(\RR)$ by the rule $\pi\cdot A=P_\pi AP_\pi^T$. The following is a classical result.
\begin{thm}
    There is a bijective correspondence between weighted digraphs having $n$ vertices and the orbits of $\mathrm{M}_{n,n}(\RR)$ under the action of the symmetric group $S_n$ by conjugation.
\end{thm}

Graph convolutions are the main building block of GNNs: they are operators that transform a graph with features into another. The vast majority of graph convolution operators $\Feat(G,n)\rightarrow\Feat(G,m)$ that are used in practice have the form we now describe. Starting from a graph with features with the choice of an ordering on the vertices $(G,F_G=(h_{v_i}\in\RR^n)_{i\in |V_G|})\in\Feat(G,n)$  the most common graph convolutions produce a graph with features $(G,F_G'=(h_{v_i}'\in\RR^m)_{i\in |V_G|})\in\Feat(G,m)$by updating the node features of the considered digraph using a \emph{message passing mechanism} that can be written in the following general form 
\begin{equation}\label{eq1}
 h_{v_i}'=\psi(h_{v_i}, \oplus_{v_j\in N^{\beta}(v_i)}\varphi(h_{v_i},h_{v_j},A_{ij},A_{ji})
 \end{equation}
where $\beta\in\lbrace h,t,u\rbrace$, $N^{h}(v_i)=\lbrace v_j\in |V_G|\: |\: A_{ji}\neq 0\rbrace$, $N^{t}(v_i)=\lbrace v_j\in |V_G|\: |\: A_{ij}\neq 0\rbrace$, $N^{u}(v_i)=N^{h}(v_i)\cup N^{t}(v_i)$, $\oplus$  denotes a permutation invariant function and $\psi$, $\varphi$ denote differentiable functions (weaker regularity assumptions can be made). Here by permutation invariant we mean that if we change the ordering chosen on the vertices (applying a permutation $\pi\in S_{|V_G|}$) we end up with the same output in the sense that 
\begin{align*}&\oplus_{v_{\pi(j)}\in N^{\beta}(v_{\pi(i)})}\varphi(h_{v_{\pi(i)}},h_{v_{\pi(j)}},(P_\pi AP_\pi^T)_{ij},(P_\pi AP_\pi^T)_{ji})= \\
&\oplus_{v_j\in N^{\beta}(v_i)}\varphi(h_{v_i},h_{v_j},A_{ij},A_{ji})
\end{align*}
In other words, these graph convolutions effectively define operators $\Feat(G,n)\rightarrow\Feat(G,m)$: the choice of an ordering on the vertices is needed in practice for the actual computations and definitions of the message passing involved but does not influence the output. They are therefore \emph{permutation equivariant}. Much of the success of the GNNs is currently being attributed to the fact that graph convolutions are \emph{permutation equivariant} in the sense we outlined above. Many popular message passing mechanisms are a particular case of the following one:
\begin{align}\label{gcneq}
    h_{v_i}'=\sum_{v_j\in N^{\beta}(v_i)}c_{f^{\beta}(i,j)}A_{f^{\beta}(i,j)}Wh_{v_j}+l_{i}A_{ii}Bh_{v_i}+e
\end{align}
here $f^{\beta}(i,j)=(i,j)$ if $v_j\in N^{t}(v_i)$ and $(j,i)$ if $v_j\in N^{h}(v_i)$, $c_{ij}$, $l_i$ are optional normalization coefficients, $W,B$ are matrices of weights and $e$ is a bias vector. For digraphs, the choice of $\beta$ should be thought of as whether a node embedding should be updated by looking at the information of the nodes that are sending to it a signal, by looking at the information of the nodes that are receiving from it a signal or both. These are three different semantics, all justifiable depending on the problem at hand. Two important graph convolutions arise as particular cases of the previous formula: Kipf and Welling's graph convolution for undirected graphs (see \cite{Kipf2016}) and GraphSage convolution (\cite{Sage}) as it appears in the popular package PyTorch Geometric (\cite{PyTorch}, \cite{PyG}.
We remark that in \cite{Kipf2016} and \cite{Sage} the above convolutions are described only for undirected graphs, but the formulas also make sense for digraphs. In fact the standard implementations used by practitioners are already capable to handle digraphs and are being used also in that context. For example, some papers introducing attention mechanisms (e.g. GAT, see \cite{GAT17}) explicitly introduce this possibility. However, in the digraph case the underlying mathematical theory is somewhat less mature (see \cite{Tong2020} and the reference therein, for the roots of the mathematics behind the directed graph Laplacian the reader is referred to \cite{Bauer2012}).
\begin{defn}\label{def:updown}
    Consider a digraph morphism $\varphi:H\rightarrow G$. We define:
    \begin{itemize}
        \item A simple upsampling from $\Feat(G,n)$ to $\Feat(H,n)$ to be the pullback $\varphi^\ast:\Feat(G,n)\rightarrow\Feat(H,n)$ associating to $(G,F_G=(h_{v})_{v\in V_G})$ the digraph with features $(H,\varphi^\ast F_G:=(h_{\varphi_V(v)})_{v\in V_H})$.
        \item If $\varphi_V$ is surjective, a simple downsampling from $\Feat(H,n)$ to $\Feat(G,n)$ to be the operator $\varphi_\ast:\Feat(H,n)\rightarrow\Feat(G,n)$ taking a graph with features $(H,F_H=\{h_v\}_{v\in V_H})$ to $(G,\varphi_\ast F_H=\{\varphi_\ast h_w\}_{w\in V_G})$ where $\varphi_\ast h_w=\sum_{v\in\varphi^{-1}(w)}h_v$.
    \end{itemize}
\end{defn}
Finally, we give the following operative definition of a graph neural network.
\begin{defn}
    Consider two weighted digraphs $G$ and $H$. A graph neural network is defined to be an operator $\Feat(G,n)\rightarrow\Feat(H,m)$ that is obtained concatenating together graph convolutions, simple graph upsampling and downsampling and activation functions.
\end{defn}
\begin{rem}
    Explicitly note that in the previous definition of GNN we allow for the convolutions and the (up)downsampling involved in a given GNN to arise as convolutions not involving necessarily the starting graph $G$ and the ending graph $H$. That is, many different intermediate graphs can be used inside a GNN. In addition, we also underline that GNNs using popular mechanisms such as skip-connections or residual blocks, fit in our definition of GNNs.
\end{rem}

\subsection{Graph Neural Networks for time series as directed graphs}
Data having the structure of a sequence can be given a structure of (weighted directed) graph with features where nodes are the \emph{words} or units in the sequence, the node features are the information associated to them and the edges are the relations between these entities. In the case of time series, each node can represent a time step, features associated to it the information about that time step and directed edges the temporal relations: for example, to preserve time coherence the graph structure can be modeled in such a way that each node can receive information only from nodes in the past. We introduce two basic examples in order to illustrate how a time series can be given a structure of digraph with features.\\
\textbf{Example 1}: A multivariate time series $\textbf{x}\in\TS(n,m)\cong\RR^{n\times m}$ can be seen as an unweighted digraph with features as follows. To start with, we consider a set of $n\times m$ nodes $v_{ij}$, with $i=1,...,n$ and $j=1,...,m$.  Then, we create edges  $v_{ij}\rightarrow v_{lk}$ only if $l\geq i$ and the edge is not a self loop (i.e. edges receive information only from the present and the past).
We assign the scalar $\textbf{x}(i-1)_j$ as feature for each node $v_{ij}$. This construction results in an unweighted digraph with features $(G_\textbf{x}, F_\textbf{x}=(\textbf{x}(i)_{j}\in\RR))$. One can modify the topology of the graph just constructed. For example, one could create edges $v_{ij}\rightarrow v_{lk}$ only if the edge is not a self-loop, $l\geq i$ and $l-i=0,1$ or if the edge is not a self-loop, $l\geq i$, and $l-i$ is both divisible by a given positive integer $d$ and is smaller than $k\cdot d$ for a given positive integer $k$. This construction results in the directed graph structure pioneered in \cite{BPabstract}): see Figure \ref{tgraph}.\\
\textbf{Example 2}: A multivariate time series $\textbf{x}\in\TS(n,m)\cong\RR^{n\times m}$ can be seen as a one dimensional time series with $m$ channels. In this case, the time series can be turned into a digraph with features $(G_\textbf{x}, F_\textbf{x}=(\textbf{x}(i-1)\in\RR^m))$ by considering a directed graph of $n$ nodes $v_{i}$, $i=1,...,n$ and edges $v_{i}\rightarrow v_{l}$  are added if the edge is not a self-loop and $l\geq i$, $l-i=1$ or and $l-i$ is both divisible by a given positive integer $d$ and is smaller than $k\cdot d$ for a given positive integer $k$. We assign the vector $\textbf{x}(i-1)\in\RR^m$ as a vector of features for each node $v_{i}$. This completes the construction of the desired directed digraph with features.\par
These examples are of course just a starting point and one can take advantage of further domain knowledge to model the topology of the graph in a more specific way. For instance, already using the basic examples provided, one could use auto-correlation functions, usually employed to determine the order of ARMA models (see \cite{TsayTS}), to choose the right value for parameters like $k$ or $d$. At this point, the reader might wonder whether certain 1D convolution can be realized as graph convolutions given a certain graph structure. This is the content of the following proposition.

\begin{prop}\label{1dconv}
Consider a convolution $\mathrm{conv1D}:\TS(r,1)\rightarrow\TS(l,1)$, $\mathrm{conv1D}(\textbf{x})(j)=\sum_{i=0}^{f-1}K_i\textbf{x}(\alpha(j,i))+b$ (as in Definition \ref{def:conv}).
Then there exist weighted digraphs $G$ and $H$ having $r,l$ nodes respectively, bijections $\psi_G :\mathrm{Feat(G,1)}\cong\TS(r,1)$, $\psi_H:\mathrm{Feat(H,1)}\cong\TS(r,1)$ and an operator $\mathrm{g}:\Feat(G,1)\rightarrow \Feat(H,1)$ that is a composition of graph convolutions of the form (\ref{gcneq}) and simple upsampling or downsampling such that $\psi_H^{-1}\circ \mathrm{conv1D}\circ \psi_G = g$.
\end{prop}
\begin{proof}
Assume that $l\leq r$ and define $G$ to be the digraph having $r$ vertices $v_1,...,v_r$ and whose weighted adjacency matrix is given by $A_{\alpha(j,i)+1,j}= K_{i}$ if $1\leq \alpha(j,i)\leq r$, $0\leq i\leq f-1$ and zero otherwise. Let $\iota:H\subseteq G$ be the weighed subgraph of $G$ consisting of the vertices $v_{1},...,v_{l}$. Define $\psi_G$ to be the map associating to $(G, h_{v_i}\in\RR)$ the time series $\textbf{x}\in \TS(r,1)$ having $\textbf{x}(i)=h_{v_{i+1}}$ and define $\psi_H$ analogously. Consider the graph convolution $\mathrm{gconv}:\Feat(G,1)\rightarrow \Feat(G,1)$ arising from the message passing mechanism given by formula (\ref{gcneq}) with $\beta = h$, $W=B=1$, $l_i=0$, $e=b$ and $c_{ij}=1$. Then $g=\iota^\ast\circ\mathrm{gconv}$. The case $l>r$ is similar.
\end{proof}
The previous lemma can be extended mutatis mutandis also to the case of multivariate time series and contains as a particular case the one of dilated convolutions. As a consequence, the process of learning the weights of a dilated convolution can be thought as the process of learning the weights of the adjacency matrix of a graph.

\begin{rem}
The previous proposition has an important conceptual consequence: ordinary TCNs can be seen as certain GNNs. As the latter are \emph{permutation equivariant} in the sense that they depend only on the graph structure and not on the ordering of the nodes, this allows us to define networks that exhibit a type of equivariance that can be leveraged: this is a desirable feature that in the case of TCNs, as we explained before (see \ref{rem:tcnlimits}), proves to be more elusive. 
\end{rem}

In addition, GNNs applied to time series can carry a different meaning than ordinary TCNs as the design of the underlying digraph employed can be modeled to exhibit other important dependencies related with the problem at hand while at the same time preserving temporal coherence. It is worth noticing that we are not suggesting that GNNs should replace TCNs or that hybrid architectures should outperform in all applications. The point of view that we are advocating is that, being the TCNs compatible with GNNs, researchers should not be constrained to sticking to one of the two frameworks only. In this sense, our work should be seen as a work laying down the theory linking TCNs and GNNs and starting to explore this link in a specific problem rather than as a work proving a thesis. For example, we do not include in our work experiments about the possible gains in terms of explainability of the hybrid digraph based models, that will be subject of future work.
\begin{rem}\label{Expltimegconv}
Simple Laplacian-based graph convolutions on undirected graphs can be seen as 1-step Euler discretizations of a heat equation (see for example \cite{Fioresi2023DeepLA}, \cite{Bronstein2016GeometricDL}). In general, graph convolutions can be thought as modeling a diffusive process of the information contained in the nodes along the edges: in our context, directed graph convolutions applied to time series digraphs \emph{diffuse} the information through time by updating at each step node features using the information coming from a \emph{temporal neighbourhood}.
\end{rem}

\section{Our Models}\label{sec:models}
To study the effectiveness of employing graph convolutions in deep learning pipelines in order to analyze time series we propose a model that takes advantage of the time-digraph structure described in the previous section: an autoencoder reconstructing the time series, that can be used for clustering or inference purposes as we shall explain later. We will first focus on the core building block of the algorithm and then proceed to discuss the model in more detail. For a more synthetic and simplified description of part of these algorithms, see also our work \cite{GSI2025}.

\begin{figure}[h]\label{fig:encdec}
\makebox[1 \textwidth][c]{
\scalebox{.7}{
\begin{subfigure}[b]{0.5\textwidth}
\centering
\begin{tikzpicture}[scale=.45]
\node at (-4,0) [inout] (input_enc) {input};
\node at (-4,-3) [main] (ly1_enc) {\begin{tabular}{c} skip-connections \\ block \end{tabular}};

\node at (-4,-6.2) [layer] (gconv_enc) {GraphConv};
\node at (-4,-8.6) [bn] (af2_enc) {$\sigma$};
\node at (-4,-11.4) [layer] (dn_enc) {Downsample};
\node at (-4,-13.6) [inout] (output_enc) {output};
\node at (-4,-17.8) {\emph{Encoder}};

\draw [->] (input_enc.south) -- (ly1_enc.north);
\draw [->] (ly1_enc.south) -- (gconv_enc.north);
\draw [->] (gconv_enc.south) -- (af2_enc.north);
\draw [->] (af2_enc.south) -- (dn_enc.north);
\draw [->] (dn_enc.south) -- (output_enc.north);

\node at (4,0) [inout] (input_dec) {input};
\node at (4,-2.2) [layer] (up_dec) {Upsample};
\node at (4,-4.4) [layer] (gconv_dec) {GraphConv};
\node at (4,-6.8) [bn] (af1_dec) {$\sigma$};
\node at (4,-10.5) [main] (ly1_dec) {\begin{tabular}{c} skip-connections \\ block \end{tabular}};
\node at (4, -13.2) [layer] (final_conv) {Final Conv};
\node at (4,-15.4) [inout] (output_dec) {output};
\node at (4, -17.8) {\emph{Decoder}};

\draw [->] (input_dec.south) -- (up_dec.north);
\draw [->] (up_dec.south) -- (gconv_dec.north);
\draw [->] (gconv_dec.south) -- (af1_dec.north);
\draw [->] (af1_dec.south) -- (ly1_dec.north);
\draw [->] (ly1_dec.south) -- (final_conv.north);
\draw [->] (final_conv.south) -- (output_dec.north);

\begin{pgfonlayer}{background}
\node at (-4,-7.5) [bg] {};
\node at (4,-5.7) [bg] {};
\node at (-1,-9.6) [times] {$\textnormal{x }m$};
\node at (7,-7.8) [times] {$\textnormal{x }m$};
\end{pgfonlayer}
\end{tikzpicture}
\caption{}
\end{subfigure}}
\hspace{1cm}
\scalebox{.7}{
\begin{subfigure}[b]{0.5\textwidth}
\centering
\begin{tikzpicture}[scale=.45]
\node at (-4,0) [inout] (input) {input};
\node at (-4,-2.2) [main] (ly1) {TCN/GCN};
\node at (-4,-4.2) [layer] (conv1) {Conv $1d$};
\node at (-4,-6.5) [bn] (af1) {$\sigma$};
\node at (-4,-8.9) [inout] (middle) {$\cdots$};
\node at (-4,-11) [main] (ly2) {TCN/GCN};
\node at (-4,-13) [layer] (conv2) {Conv $1d$};
\node at (-4,-15.2) [bn] (af2) {$\sigma$};
\node at (-4,-17.8) [inout] (output) {output};

\draw [decorate, decoration={brace,amplitude=5pt}] (0,-5) -- (0,-13);

\draw [->] (input.south) -- (ly1.north) ;
\draw [->] (ly1.south) -- (conv1.north);
\draw [->] (conv1.south) -- (af1.north);
\draw [->] (af1.south) -- (middle.north);
\draw [rounded corners=5pt,->] (af1.west) --++(-4,0) |- (output.190);
\draw [->] (middle.south) -- (ly2.north);
\draw [rounded corners=5pt,->] (middle.west) --++(-3,0) |- (output.180);
\draw [->] (ly2.south) -- (conv2.north);
\draw [->] (conv2.south) -- (af2.north);
\draw [->] (af2.south) -- (output.north);
\draw [rounded corners=5pt,->] (af2.west) --++(-3,0) |- (output.170);

\begin{pgfonlayer}{background}
\node at (-4,-4.5) [bg3] {};
\node at (-4,-13.2) [bg3] {};
\node at (2.3,-9) [times] {$n\textnormal{ times}$};
\end{pgfonlayer}
\end{tikzpicture}
\caption{}
\end{subfigure}}}
\caption{\footnotesize{The encoder and decoder (a), and the skip-connections block (b)}}\label{mb}
\end{figure}
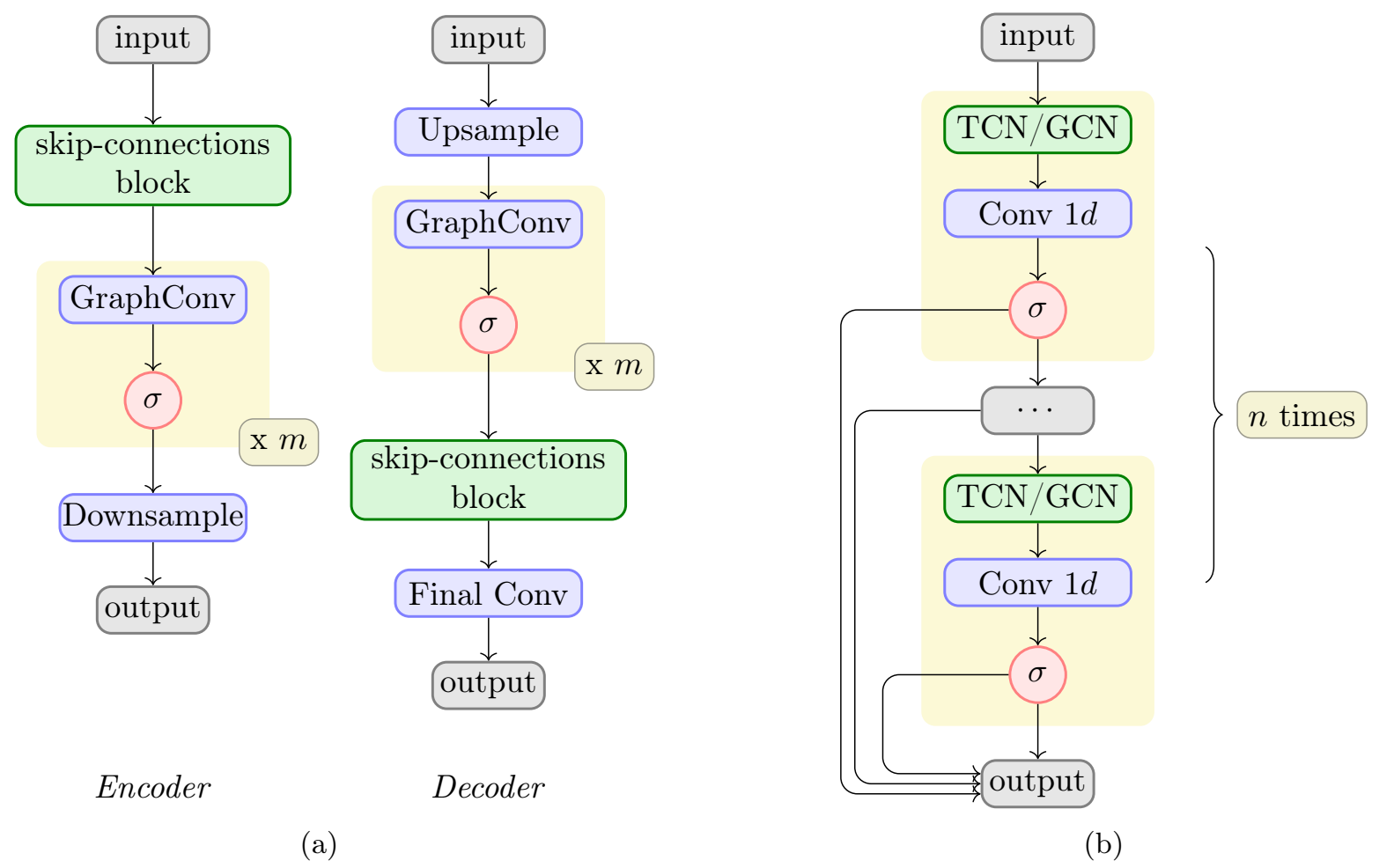

\subsection{Main Building Block}\label{subs:mainbb}
The main building block for all the models presented in this paper is a collection of layers that combines TCNs with GNNs. Inspired by what has already been done in the literature, we propose this main building block in two versions: an encoder and a decoder (they are sketched in Figure \ref{mb}(a)). They can be used on their own or put together to construct an autoencoder model.
\par
In the encoder version, the input is first given to a block of either $n$ TCN or $n$ GNN (we tested primarily Sage convolution-based layers as they appeared more effective after some preliminary tries) with skip connections, following the approach proposed in \cite{Thill2021}. In the architectures we have tested, the TCN/GNN layers are simply a temporal convolution or a graph convolution (followed by dropout and an activation function), but more involved designs are possible. The effectiveness of skip-connections in the model developed in \emph{op.cit} is due to the fact that stacking together the outputs of dilated convolutions with different dilations allows to consider short and long term time dependencies at the same time. Skip connections have been used also in the context of GNNs: for example, Sage convolutions \cite{Sage} introduce them already in the message passing mechanism. This motivates the introduction of skip connections in GNNs handling digraphs with features coming from time series: in this context they do not only allow to bundle together information coming from "temporal neighbourhoods" of different radii as in the architecture developed in \cite{Thill2021} but they also help reducing oversmoothing that traditionally curses GNNs architectures (see \cite{Bodnar2022NeuralSD} for a discussion and the references therein).
The skip-connections block is described in Figure \ref{mb}(b) and works as follows. The input goes through the first TCN/GNN layer, followed by a 1-dimensional convolution and an activation function. We call the dimension of the output of this convolution skip dimension. Then the output is both stored and passed to the next TCN/GNN layer as input and so on. At the end all stored tensors are stacked together and passed to a series of $m$ graph convolutions, each one followed by an activation function. We tested the convolutions: GCN (cfr. \cite{Kipf2016}), Sage (cfr. \cite{Sage}), GAT (cfr. \cite{GAT17}).
\par
The graph convolutions are defined to encode in the embedding of a given node, at each pass, the information provided by all the nodes in its neighbourhood. Now, looking at how a time-digraph is defined, one sees that in our set up this translates to building the embedding of a given node, that is a data point of the time series, using the information given by the data points that are close in time (short-term memory behavior) or at a certain distance $d$ away in time (long-term memory behavior), where $d$ is set in the construction of the graph.
\par
Finally the intermediate output produced by the graph convolutions is given to an optional 1d convolution with kernel size $1$ to adjust the number of channels and then to either an average pooling layer (i.e. a simple downsampling as in Definition \ref{def:updown}) or a max pooling layer (that we can see as a nonlinear simple downsampling) that shrinks the temporal dimension of the graph. In other words, if we think about the time-graph as a time window of length $T$, the pooling layers outputs a time window, and therefore a time-graph, of length $T/s$, thus realizing the characteristic \emph{bottleneck} of an autoencoder as described for instance in, \cite{Thill2021}, \cite{Khandual2021}. We will refer to $s$ as the shrinking factor.
\par
The decoder changes the order of the blocks we just described, in a symmetric way. It starts with an upsample (i.e. a simple outsample as in Definition \ref{def:updown} or a nonlinear version of it) of the input time series which is then passed to the graph convolutions followed by the skip-connections block. It terminates with a 1d final convolution with kernel size $1$ that reduces the number of channels, or hidden dimensions, thus giving as output a time series of the same dimensions as the initial input. In the case where the skip-connections block is built with GNN layers, this final convolution can be replaced by a final TCN layer. Both the encoder and the decoder can be modified to be \emph{bidirectional}. In this variant each skip-connections block is paired with an additional skip connection block having the same structure but receiving as an input the original time series reversed or if the input is a digraph with features, the digraph with features obtained from the starting digraph and having as weighted adjacency matrix the transposed of the original one. The outputs of these two parallel blocks is then summed together and fed to the following block as before. With this modification, our model mimics bidirectional architectures sometimes found in RNN-based models (\cite{BiRNN}).

\subsection{Autoencoders and Variational Autoencoders}
\begin{figure}[h]
\begin{center}
\scalebox{.90}{
\begin{tikzpicture}[scale=.45]
\node at (-3,0) [inout] (input) {$G_i$};
\node at (-3,-2.5) [main] (enc) {Encoder};
\node at (-3,-4.5) [main] (dec) {Decoder};
\node at (3,-2.5) [layer] (mlp) {MLP};
\node at (3,-6.5) [inout] (out_hdcl) {label};
\node at (-3,-6.5) [inout] (output) {$\hat G_i$};

\draw [->] (input.south) -- (enc.north);
\draw [->] (enc.south) -- (dec.north);
\draw [->] (dec.south) -- (output.north);
\draw [dashed,->] (enc.east) -- (mlp.west);
\draw [dashed,->] (mlp.south) -- (out_hdcl.north);

\end{tikzpicture}}
\end{center}
\caption{\footnotesize{The structure of the autoencoder.}}\label{aetcn}
\end{figure}
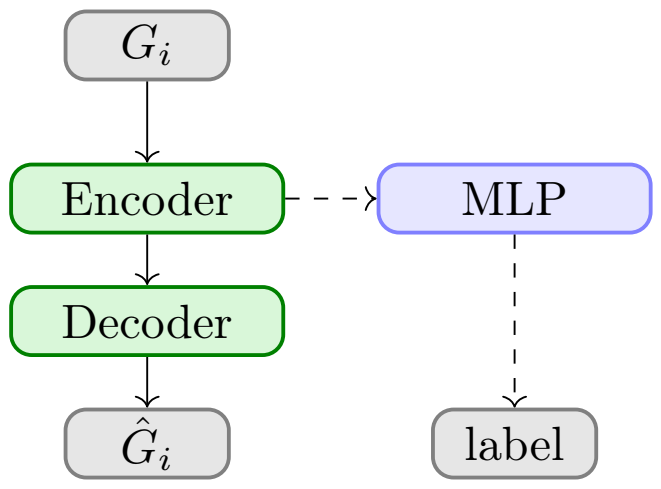
We propose an autoencoder model, together with its variational counterpart. The autoencoder employs two main building blocks as in figure \ref{mb}(a), first used as an encoder and then as a decoder and it produces as output the reconstruction of the given time series represented by the time graph. This is the architecture we employed in \cite{GSI2025}. This model might include the following modification: after the encoding, the tensors in the hidden layer are passed both to the encoder and to a multi-layer perceptron classifier whose output is a label $\hat{y}$ (see Figure \ref{aetcn}). The output $\hat{y}$ is confronted with the true label of the classification task using a suitable loss function $\mathcal{L}_{\textnormal{hdcl}}$ and concurs to the total loss computation, together with the reconstruction loss $\mathcal{L}_{\textnormal{ae}}$:
\begin{equation}\mathcal{L}_{\textnormal{tot}} = \mathcal{L}_{\textnormal{ae}} + \mathcal{L}_{\textnormal{hdcl}}\end{equation}

The variational autoencoder is built using exactly the same encoder and decoder, as usual. As for the loss, it is composed of the following contributions (see \cite{infovae, vaeKW}).
\begin{equation}\mathcal{L}_{\textnormal{tot}} = \mathcal{L}_{\textnormal{ae}} + \mathcal{L}_{\textnormal{mmd}} + \mathcal{L}_{\textnormal{hdcl}}\end{equation}
where the term $\mathcal{L}_{\textnormal{hdcl}}$ is optional, as for the case of the standard autoencoder, while $\mathcal{L}_{\textnormal{mmd}}$ is the maximum mean discrepancy \cite{gretton, infovae}. As explained in \emph{op. cit.} divergence measures how similar two distributions are by comparing all of their moments, using a kernel embedding trick. If $p(x)$ and $q(x)$ are the two distributions and $k(x,x')$ is a positive definite kernel, then the MMD divergence is defined as follows:
\begin{equation}
    \mathcal{L}_{\textnormal{mmd}}=\mathbb{E}_{p(x),p(x')}[k(x,x')] + \mathbb{E}_{q(x),q(x')}[k(x,x')] - 2\mathbb{E}_{p(x),q(x')}[k(x,x')]
\end{equation}
Roughly, the more the two distributions are similar, the lower $\mathcal{L}_{\textnormal{mmd}}$ will be, as the similarity of samples from each distribution should be close to the similarity of samples takes from both distributions, in other words $\mathcal{L}_{\textnormal{mmd}}=0$ if and only if $p=q$.
For our experiments we use the Gaussian kernel $k(x,x')=e^{-\frac{||x-x'||^2}{2\sigma^2}}$.

From now on we will refer to the AE models (respectively to the VAE models) as:
TCNGraphAE (TCNGraphVAE) if the skip-connections block uses TCN layers, TGraphAE (TCNGraphVAE) if it is built with graph convolutions and TGraphMixedAE (TCNGraphVAE) if the encoder uses graph convolutions and the decoder uses TCN layers. We will add the suffix hdcl if the total loss function of either the autoencoder or the variational autoencoder includes the term $\mathcal{L}_{\textnormal{hdcl}}$. If the skip connections blocks consist only of dilated convolutions and we do not have a final graph convolution to filter the signal, we obtain a TCN autoencoder with a structure similar to the one described in \cite{Thill2021}. We call this model TCNAE and regard it as the state of the art model in this context, using it as a benchmark.

\section{Experiments and results}\label{sec:exp}
We test our algorithms to solve anomaly detection problems concerning ECG datasets. To this aim we used the benchmark dataset called ECG5000, that is also considered by many other authors that developed and tested their algorithms on it. We also detail in Section \ref{sec:indval} the results we obtained on an industry validation set we considered in \cite{GSI2025}. As the pipeline proposed in this paper has been tested on the ECG5000 dataset only, further experiments would be needed to confirm its efficacy on different datasets. While the ECG5000 can be considered as one of the main benchmark datasets it has limits (see \cite{ECGNET}).

\subsection{Dataset ECG5000}

The ECG5000 dataset is a benchmark dataset comprising of 5000 ECG signals that has been used by many authors (see \cite{ECGNET} for a comprehensive introduction to the literature). 
Signals are labeled as normal or as anomalous depending on whether or not they represent normal rhythms. We did not preprocess these signals. As for the split, there is no consensus among the authors about which is the most appropriate split. Accordingly, we performed a train/valid/test split that we now explain. As a train set, we kept the first 500 signals of the database: this is the same training set that the authors considering only 500 samples as their training set consider. In the training set $58\%$ of samples were normal, for a total of $292$ normal samples. From the remaining samples we selected randomly 256 signals to form our valid set. The signals of the dataset not used in the train set and in the valid set have been used as a test set. The percentage of normal signals in the validation set was $57\%$ and $58\%$ in the test set ($147$ and $2480$ normal samples respectively). Accordingly our split resulted in essentially equally balanced sub-datasets. We adopted a strategy different from the one we employed in the case of the signal quality recognition. We decided to follow the approach adopted in \cite{Pereira2019}, for example, that is the one to use an autoencoder to embed the signals into a latent space where a clustering or a classification is performed. To help modeling the latent space in such a way that distinguishing the normal from the abnormal signals becomes easier, we decided to employ autoencoders feeding the hidden representations of the signals to an MLP for classification, as explained in Section \ref{sec:models}. Each model we considered has been used as follows:
\begin{itemize}
    \item[1)] The model is trained on the training set: here, to the time series representing the signals the same structure of directed one used for the industry dataset has been employed for the graph convolutions involved in the model. To each time series $\textbf{x}$, a directed graph structure was given as described in Example 1. In particular, each time step $\textbf{x}_i$ is a node and we have a directed edge $\textbf{x}_k\rightarrow \textbf{x}_i$ for each $k=i-1,i-d,i-2d,...,i-n\cdots$ with $d=8$. The number $n\cdot d$, called \emph{lookback window}, was set to be $\frac{1}{5}$ of the total length of our sequences. All the graph convolutions appearing in the models were implemented in their directed version: the resulting convolutions then updated the node features of a node using only the information of nodes in the past. We chose $d=8$ as the majority of the GNN skip-connection blocks we used had $7$ layers. Combined with our lookback window size, this depth enables each point in the block's output to aggregate information across the entire time series while maintaining a broader, more diluted receptive field. This lookback window choice was specifically designed to allow graph convolutions to cover the full sequence length twice. Furthermore, dividing the sequence into $5$ sub-segments aligns naturally with standard ECG morphology, as a single heartbeat is broadly characterized by three major phases: the P-wave, QRS complex, and T-wave (corresponding to atrial depolarization, ventricular depolarization, and ventricular re-polarization, respectively). Between the first and the second the so called PR segment can be found and between the second and the third phase usually the ST segment is identified amounting to roughly 5 sub-intervals of an ECG waveform. The reader should think of this as a heuristic: we did not pinpoint the exact start and end points of these subdivisions. However, we believe that this choice can serve as a first default choice of this parameter. This choice is also aligned with the one we did in our work \cite{GSI2025}.
    \item[2)] An SVC classifier is trained on the hidden representations of the signals in the latent space. Also, a K-means algorithm is applied to the representations in the latent space of the model to find two centroids that can be used for binary classification.
    \item[3)] Classification is performed on the valid set using the SVC and the centroids obtained in the previous point. In addition, the MLP classifier trained in the training is also used on the valid set to evaluate its performance and a K-means algorithm is trained from scratch on the valid set to cluster the hidden representations into two clusters.
\end{itemize}

\begin{table*}[ht]
\begin{center}
\footnotesize
\scalebox{.84}{
\makebox[1 \textwidth][c]{
\begin{tabular}{c | c c c | c | c}
\textbf{Model} & \textbf{Precision} & \textbf{Recall} & \textbf{F1 Score} &  \textbf{Accuracy} & \textbf{Parameters}\\
\hline
TGraphAE\_hdcl & $0.974 \pm 0.005$ & $0.973 \pm 0.004$ & $0.975 \pm 0.002$ & $0.979 \pm 0.001$ & $46 k$ \\
TGraphMixedAE\_hdcl & $0.979 \pm 0.005$ & $0.968 \pm 0.008$ & $0.974 \pm 0.004$ & $0.979 \pm 0.001$ & $158 k$ \\
TCNGraphAE1\_hdcl & $0.985 \pm 0.004$ & $0.978 \pm 0.007$ & $0.981 \pm 0.001$ & $0.985 \pm 0.001$ & $287k$ \\
TCNGraphAE2\_hdcl  & $\textcolor{purple}{0.991 \pm 0.001}$ & $0.978 \pm 0.003$ & $\textcolor{purple}{0.984 \pm 0.001}$ & $\textcolor{purple}{0.987 \pm 0.001}$ & $159 k$ \\
TCNGraphAE3\_hdcl  & $0.983 \pm 0.007$ & $0.978 \pm 0.003$ & $0.981 \pm 0.002$ & $0.984 \pm 0.002$ & $43 k$ \\
TCNGraphAE4\_hdcl & $0.984 \pm 0.003$ & $\textcolor{purple}{0.990 \pm 0.004}$ & $0.982 \pm 0.003$ & $0.985 \pm 0.003$ & $82 k$ \\
\hline
\end{tabular}}}
\end{center}\caption{\footnotesize{Results of the standard autoencoders for label 0. Best scores are colored in purple.}}\label{tabae}
\end{table*}

\begin{table*}[ht]
\begin{center}
\footnotesize
\scalebox{.84}{
\makebox[1 \textwidth][c]{
\begin{tabular}{c | c c c | c | c}
\textbf{Model} & \textbf{Precision} & \textbf{Recall} & \textbf{F1 Score} &  \textbf{Accuracy} & \textbf{Parameters}\\
\hline
TGraphAE\_hdcl & $0.980 \pm 0.003$ & $0.982 \pm 0.004$ & $0.982 \pm 0.002$ & $0.979 \pm 0.001$ & $46 k$ \\
TGraphMixedAE\_hdcl & $0.979 \pm 0.002$ & $0.985 \pm 0.003$ & $0.982 \pm 0.001$ & $0.979 \pm 0.001$ & $158 k$ \\
TCNGraphAE1\_hdcl & $0.984 \pm 0.005$ & $0.990 \pm 0.003$ & $0.987 \pm 0.001$ & $0.985 \pm 0.001$ & $287k$ \\
TCNGraphAE2\_hdcl  & $\textcolor{purple}{0.985 \pm 0.002}$ & $0.994 \pm 0.001$ & $\textcolor{purple}{0.989 \pm 0.001}$ & $\textcolor{purple}{0.987 \pm 0.001}$ & $159 k$ \\
TCNGraphAE3\_hdcl  & $0.985 \pm 0.002$ & $0.988 \pm 0.005$ & $0.987 \pm 0.002$ & $0.984 \pm 0.002$ & $43 k$ \\
TCNGraphAE4\_hdcl & $0.980 \pm 0.005$ & $\textcolor{purple}{0.995 \pm 0.001}$ & $0.988 \pm 0.002$ & $0.985 \pm 0.003$ & $82 k$ \\
\hline
\end{tabular}}}
\end{center}\caption{\footnotesize{Results of the standard autoencoders for label 1. Best scores are colored in purple.}}\label{tabae1}
\end{table*}

We now clarify the relations between the components of our pipeline. The MLP classifier is trained jointly with the autoencoder: as the latent representations of the signals are exposed to class labels via the MLP, the autoencoder's latent space is shaped accordingly. The SVC and K-means algorithms are subsequently trained independently on the latent representations extracted by the final trained model. 
This provides us with four distinct strategies to assign labels to a new batch of data:

\begin{enumerate}[label=\alph*)]
    \item The prediction from the trained MLP evaluated on the latent representation of the signal.
    \item The prediction from the trained SVC evaluated on the latent representation of the signal.
    \item The pseudo-label assigned by the K-means model (trained on the training set's latent representations) evaluated on the latent representation of the signal.
    \item The pseudo-label assigned by a new K-means model retrained directly on the latent representations of the new data batch.
\end{enumerate}

Methods a) and b) operate in a fully supervised manner. In cases c) and d), an unsupervised algorithm, K-means, is used. However, because the latent embeddings were shaped by joint training with a supervised MLP, we classify these approaches as semi-supervised rather than strictly unsupervised.
We performed a grid-search to test several models and different parameters repeating for each choice the 3 steps above and comparing the results obtained by the labelings a)-d). For each different model, we repeated this process 3 times using different random seeds. Finally, we selected, for each model type, the best performing models for evaluation on the test set. The evaluation on the test set has been conducted as in points 1)-3) above using the test set in place of the valid set: in this case we trained and evaluated our models for 10 times with 10 different random seeds, we discarded the best and the worst results for each score considered and then we computed the mean and standard deviation of the remaining scores.\\ Interestingly, the method that achieved an higher performance on both the valid and the test set was method c), similar to the performance of method d). As the former can be in principle used to label single out of sample signals we then reported in our paper the results obtained with this method. This behavior appeared to be consistent across essentially all good performing models. We report the results obtained with standard autoencoders in Table \ref{tabae} and Table \ref{tabae1} while we report the results obtained using the variational autoencoders in Table \ref{tabvae} and in Table \ref{tabvae1}. Benchmark results are listed in Table \ref{tabbench}: we have selected them by looking at results tabulated for models trained using train/valid/test splits similar to the ones we used in the paper \cite{AttentivECGRU} that we believe to be the most up to date on this particular problem.

\begin{table*}[t]
\begin{center}
\footnotesize
\scalebox{.84}{
\makebox[1 \textwidth][c]{
\begin{tabular}{c | c c c | c | c}
\textbf{Model} & \textbf{Precision} & \textbf{Recall} & \textbf{F1 Score} &  \textbf{Accuracy} & \textbf{Parameters}\\
\hline
TGraphVAE\_hdcl & $0.979 \pm 0.004$ & $0.976 \pm 0.003$ & $0.976 \pm 0.004$ & $0.980 \pm 0.003$ & $182k$ \\
TGraphMixedVAE\_hdcl & $0.977 \pm 0.003$ & $0.975 \pm 0.002$ & $0.976 \pm 0.002$ & $0.980 \pm 0.002$ & $145k$ \\
TCNGraphVAE1\_hdcl & $0.992 \pm 0.002$ & $\textcolor{purple}{0.977 \pm 0.004}$ & $\textcolor{purple}{0.984 \pm 0.002}$ & $\textcolor{purple}{0.987 \pm 0.001}$ & $178k$ \\
TCNGraphVAE2\_hdcl & $\textcolor{purple}{0.993 \pm 0.001}$ & $0.972 \pm 0.003$ & $0.983 \pm 0.001$ & $0.986 \pm 0.001$ & $322k$ \\
\hline
\end{tabular}}}
\end{center}\caption{\footnotesize{Results of the variational autoencoders for label 0. Best scores are colored in purple.}}\label{tabvae}
\end{table*}

\begin{table*}[t]
\begin{center}
\footnotesize
\scalebox{.84}{
\makebox[1 \textwidth][c]{
\begin{tabular}{c | c c c | c | c}
\textbf{Model} & \textbf{Precision} & \textbf{Recall} & \textbf{F1 Score} &  \textbf{Accuracy} & \textbf{Parameters}\\
\hline
TGraphVAE\_hdcl & $0.983 \pm 0.002$ & $0.985 \pm 0.003$ & $0.983 \pm 0.003$ & $0.980 \pm 0.003$ & $182k$ \\
TGraphMixedVAE\_hdcl & $0.982 \pm 0.002$ & $0.984 \pm 0.002$ & $0.983 \pm 0.002$ & $0.980 \pm 0.002$ & $145k$ \\
TCNGraphVAE1\_hdcl & $\textcolor{purple}{0.984 \pm 0.002}$ & $0.994 \pm 0.001$ & $\textcolor{purple}{0.989 \pm 0.001}$ & $\textcolor{purple}{0.987 \pm 0.001}$ & $178k$ \\
TCNGraphVAE2\_hdcl & $0.980 \pm 0.002$ & $\textcolor{purple}{0.995 \pm 0.001}$ & $0.988 \pm 0.001$ & $0.986 \pm 0.001$ & $322k$ \\
\hline
\end{tabular}}}
\end{center}\caption{\footnotesize{Results of the variational autoencoders for label 1. Best scores are colored in purple.}}\label{tabvae1}
\end{table*}


\begin{figure*}[t]
\begin{subfigure}[b]{0.45\textwidth}
\includegraphics[scale=0.3]{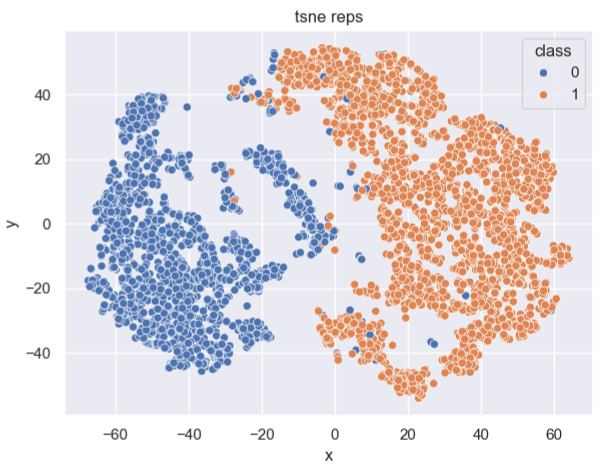}
\caption{}
\end{subfigure}
\begin{subfigure}[b]{0.45\textwidth}
\includegraphics[scale=0.3]{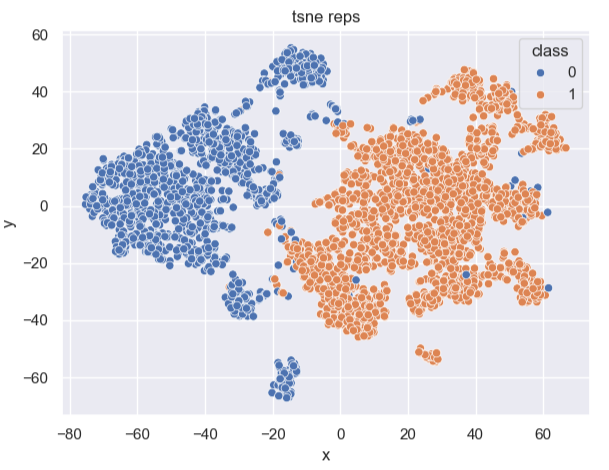}
\caption{}
\end{subfigure}
\caption{\footnotesize{Representations of the hidden embeddings of the signals obtained by TGraphMixedVAE\_hdcl (a) and TCNGraphVAE1\_hdcl (b), and converted to 2D representations using TSNE. Label $0$ indicates a normal signal, $1$ an abnormal signal.}}
\label{figclustecg5000}\end{figure*}

\subsubsection{Discussion}

\begin{table*}[ht]
\begin{center}
\footnotesize
\scalebox{.84}{
\makebox[1 \textwidth][c]{
\begin{tabular}{m{2.2cm}|c|c|c|c|c|p{1.2cm}}
\textbf{Model} & \textbf{Train/Val/Test} & \textbf{Accuracy} & \textbf{Precision} & \textbf{Recall} & \textbf{F1 Score} & \textbf{Params} \\
\hline
Proposed best,\newline TCNGraphVAE1 & 0.1/0.05/0.85 & 98.7\%$\pm$ 0.1 & 0.984$\pm$ 0.002 & 0.994$\pm$ 0.001 & 0.989$\pm$ 0.001 & $178k$ \\
\hline
Proposed best small,\newline TCNGraphAE4 & 0.1/0.05/0.85 & 98.5\%$\pm$ 0.3 & 0.980$\pm$ 0.005 & 0.995$\pm$ 0.001 & 0.988$\pm$ 0.002 & \textcolor{purple}{$82k$}\\
\hline
\hline
Pereira et al. \cite{Pereira2019},\newline VAE + SVM & 0.08/0.02/0.9 & 98.43\% & 0.9847 & 0.9843 & 0.9844 &  $273k$\\
\hline
Matias et al. \cite{Matias2021},\newline VAE + local similarity score & 0.05/0.05/0.9 & 97.11\% & – & – & 96.01 & $250k$, \newline (estimate) \\
\hline
\hline
ECG-NET \cite{ECGNET},\newline LSTM AE & 0.05/0.005/0.9 & 98.56\% & 0.9794 & \textcolor{purple}{0.9957} & 0.9875 &  $1649k$ \\
\hline
AttentivECGRU \cite{AttentivECGRU}\newline GRU AE+attention & 0.05/0.005/0.9 & \textcolor{purple}{98.96\% } & \textcolor{purple}{0.9912} & 0.9908 & \textcolor{purple}{0.9909} & $743k$ \\
\hline
\end{tabular}}}
\end{center}\caption{\footnotesize{Comparison of our results with other existing methods applied on the ECG5000 dataset. Best scores are colored in purple. The final two models reported in the table use only the normal samples in their train/valid sets. However, note that the ordering of the dataset has been preserved in all experiments: as a consequence their training and validation sets are subsets of our training set and our test set is a subset of theirs.}}\label{tabbench}
\end{table*}

From the experimental results it appears that in this case the best autoencoder architecture has an encoder and a decoder as detailed in Figure \ref{mb}. Also, while Variational autoencoders proved on average to perform better when they had enough parameters, we noticed that the ordinary ones were able to achieve a good performance also in with model having a reduced number of parameters. We deem the number of parameters to be an important metric to evaluate the effectiveness of an algorithm: a small model can be deployed on more devices after appropriate quantization. From Table \ref{tabbench} it emerges that our proposed best model, a variational graph-based autoencoder, has a performance superior to the ones of the other variational autoencoders considered in our comparison, while at the same time having a smaller number of parameters. In addition, our proposed best small model still has high scores in the other metrics considered. This is noteworthy as this model has roughly half of the parameters of our proposed best model and roughly a tenth of the parameters of the model proposed in \cite{AttentivECGRU} that, to our knowledge, is the one achieving the best accuracy overall. In addition, while the model developed in \cite{AttentivECGRU} labels the signals using an autoencoder's reconstruction error, ours uses the embedding into the latent space of an autoencoder to assign to each signal a label: as a result our algorithm, when deployed, has roughly half the size of our full trained model (as we need only the encoder of our autoencoder). In addition, the models we proposed appear to be stable under change of the seed used for the training: unfortunately, we do not know if the results reported by the other models we report in Table \ref{tabbench} have a similar behavior or if in \emph{op. cit.} only one seed has been considered. Finally we report in table \ref{tab:paired_wilcoxon_accuracy} the Wilcoxon signed-rank test across random seeds for selected pairs of models. It emerges that there is no statistically significant difference among models with a similar encoder and decoder structure while a significant one emerges when considering the comparisons between the models only involving graphs (TGraphAE/VAE) with the ones having TCNs both in the encoder and in the decoder (TCNGraphAE/VAE). 

\begin{table*}[t]
\begin{center}
\small
\scalebox{.84}{
\makebox[1 \textwidth][c]{

\begin{tabular}{l|l|l|p{4cm}}
\textbf{Model A} & \textbf{Model B} & \textbf{Wilcoxon} $p$-value & \textbf{Main difference} \\
\hline
TCNGraphAE4\_hdcl &
TCNGraphAE3\_hdcl &
0.5566 & Bidirectional and not bidirectional \\

TCNGraphAE1\_hdcl &
TCNGraphAE4\_hdcl &
0.9527 & Lowest number of parameters vs greatest\\

TCNGraphVAE1\_hdcl &
TCNGraphAE4\_hdcl &
0.3750 & Best variational vs best non-variational\\

TGraphMixedAE\_hdcl &
TGraphAE\_hdcl &
0.3726 & Different decoder structure \\

TGraphMixedVAE\_hdcl &
TGraphVAE\_hdcl &
1.0000 & Different decoder structure\\

TCNGraphVAE1\_hdcl &
TGraphVAE\_hdcl &
0.0020 & Different encoder and decoder structure \\

TCNGraphAE4\_hdcl &
TGraphAE\_hdcl &
0.0020 & Different encoder and decoder structure \\
\hline
\end{tabular}}}
\end{center}
\caption{Paired Wilcoxon signed-rank tests across ten training seeds. The final two comparisons are statistically significant.}\label{tab:paired_wilcoxon_accuracy}
\end{table*}

\subsection{Validation on industry data}\label{sec:indval}
In our preliminary work \cite{GSI2025} we used unsupervised, non-variational variants of the models we employed in this paper to solve a quality detection problem. In that work, we applied a fully unsupervised method on an industry dataset of signals recorded using the ButterfLife Device \cite{BF} and that was provided to us by the company VST. The results we obtained confirmed that graph-based TCNs are competitive and can surpass state-of-art TCN-based architectures for these types of problems, while allowing for the development of models having a restricted number of parameters. We reproduce in appendix \ref{app:qdect} the table of the results obtained in \emph{op. cit.}

\section{Conclusions}

In this paper, we explained how to see TCNs as particular GNNs and we explained why from a theoretical point of view it might make more sense to consider time series as directed graphs for the design of deep learning architectures. This can be seen as one of the main goals of the paper: to introduce a theoretical framework justifying the combined usage of TCNs and GNNs. Our experiments confirmed that, if the cases we considered, graph convolutional layers added to established algorithms can help to reduce their number of parameters while not decreasing their performance and adding a slight stabilizing effect. As a third and final main contribution, we introduced a specific semi-supervised, parameter-efficient pipeline for the study of ECG signals. An important point is that only the encoder of the trained autoencoders used in the pipeline and the K-means centroids coordinates are needed after training to label signals in inference, thus reducing roughly by half the resources needed to use the trained autoencoder: this contrasts with other approaches such as the one of \cite{ECGNET} that are based on the reconstruction error of an autoencoder: in these cases running the full autoencoder is necessary for inference and labeling purposes. We plan to continue the study of GNNs applied to time digraphs with features in the context of multivariate time series, constructing more complex time digraph structures and using more capable message passing mechanisms.
\section{Funding Details}
The research of the first author has been supported by the project PRIN - FRAIL and by  CaLISTA CA 21109. The research of the second author has been supported by CaLISTA CA 21109, CaLIGOLA MSCA-2021-SE-01-101086123, MSCA-DN CaLiForNIA—101119552, PNRR MNESYS, the PNRR National Center for HPC, Big Data, and Quantum Computing CUP J33C22001170001, SimQuSec CUP J13C2200068000; INFN Sezione Bologna, Gast initiative and GNSAGA Indam.
\section{Author Contribution}
All authors contributed equally to this work.
\section{Data availability statement}
The description of the ECG5000 dataset can be found at \url{https://www.timeseriesclassification.com/description.php?Dataset=ECG5000}; we downloaded it from the TensorFlow repository \url{http://storage.googleapis.com/download.tensorflow.org/data/ecg.csv}. The dataset as it was downloaded is also available as a csv file in the supplementary materials of this paper. The source code is available upon request.

\appendix

\section{Models hyperparameters and implementation}\label{hyp}
We trained our model on workstations having RTX4090 and RTX5090 GPUs and we used the PyTorch library to write our models. All the remaining code was written in python and used standard numerical libraries such as Numpy, Pandas and Scikit-learn. To train our models, we used the Adam optimizer with a learning rate of 0.001 and the parameter \texttt{drop\_last=True}. To make our trainings reproducible we used built-in PyTorch methods to seed the random number generators called in the trainings we performed. Source code is available in the public repository \emph{link added upon acceptance, until then we prefer it to stay upon request}.
We detail here all the hyperparameters used in the training of our models. We list the most significant ones in two distinct tables: they correspond exactly to the models presented in the main body of the paper and whose results are presented in Table \ref{tabvae1}. We write here a short description and recollection of all the terms used in these tables:
\begin{itemize}
    \item \texttt{Num Channels}: the number of skip connection layers and their output dimensions (list). The value after the comma is the bottleneck channel dimension post-encoder's 1D graph convolution. The dimension used in the decoder are the same but in inverse order, except that in the case of the model flagged with the symbol $\diamond$.
    \item \texttt{Skip dims}: skip dimensions as per Section \ref{sec:models}.
    \item \texttt{GConv type}: graph convolution details (if present) following the skip block: convolution type and hidden dimensions. For GAT, it includes the number of heads (e.g., GAT2H for 2 heads).
    \item \texttt{Pool type}, \texttt{Downsample}, \texttt{Upsample}: standard layer notation is used (e.g., PyTorch Geometric).
    \item \texttt{Shrink}: the reduction factor of the average pooling applied by the downsample at the end of the encoder.
    \item \texttt{Kernel sizes}: a list of the kernel sizes of each 1D convolution employed.
    \item \texttt{Dilations} a list of the dilation factor of each 1D convolution employed.
\end{itemize}

Tables \ref{tab:params_ae} and \ref{tab:params_vae} list all model hyperparameters, found using a grid search, for the dataset ECG5000.
We applied \texttt{SiLU} activation, batch normalization, and dropout. The models TGraphAE\_hdcl and TGraphVAE\_hdcl feature a final 1D dilated convolution (dilation 1) after their decoder. The models TGraphMixedAE\_hdcl and TGraphMixedVAE\_hdcl are listed with an asterisk to signal the fact that in this case the hyperparameters \texttt{num Channels} and \texttt{skip dims} are not the same for the encoder and the decoder. The ones reported in the table are those of the encoders, that are made of graph convolutions. The decoders had twice the number of \texttt{num Channels} and \texttt{skip dims} of the correspondent encoders. Finally, the models marked with a dagger $\dagger$ are bidirectional (see Section \ref{sec:models}).
\begin{table*}[t!]
\begin{center}
\footnotesize
\scalebox{.75}{
\makebox[0.8\textwidth][c]{
\begin{tabular}{c | c c c c c c }
\multicolumn{7}{c}{Autoencoders}\\
\hline
\textbf{Model}  & \textbf{Num Channels} & \textbf{Skip dims} & \textbf{GConv type} & \textbf{Kernel sizes} & \textbf{Dilations} & \textbf{Shrink} \\
\hline
TGraphAE\_hdcl & $[128, 64, 32, 16, 8, 4, 2]$ & $[64, 32, 16, 8, 4, 2]$ & Sage & - & - & $35$\\
TGraphMixedAE\_hdcl* & $[64]*7,8$ & $[16]*7$ & Sage, $[100]$ & $[3]*7$ & $[1, 2, 4, 8, 16, 32, 64]$ & $20$ \\
TCNGraphAE1\_hdcl & $[128, 112, 96, 80, 64, 48, 8]$ & $[64, 48, 32, 16, 8, 4]$ & GAT1H & $[8]*6$ & $[1, 2, 4, 8, 16, 32]$ & $35$\\
TCNGraphAE2\_hdcl$^\dagger\diamond$ & $[64, 48, 32, 16, 8, 4, 4]$ & $[32, 16, 8, 4, 2, 2]$ & Sage, $[16, 16]$ & $[8]*6$ & $[1, 2, 1, 4, 1, 8]$ & $20$\\
TCNGraphAE3\_hdcl & $[64, 48, 32, 16, 8, 4, 4]$ & $[32, 16, 8, 4, 2, 2]$ & Sage, $[16, 16]$ & $[8]*6$ & $[1, 2, 1, 4, 1, 8]$ & $10$\\
TCNGraphAE4\_hdcl$^\dagger$ & $[64, 48, 32, 16, 8, 4, 4]$ & $[32, 16, 8, 4, 2, 2]$ & Sage, $[16, 16]$ & $[8]*6$ & $[1, 2, 1, 4, 1, 8]$ & $10$\\
\hline
\end{tabular}}}
\end{center}
\caption{\footnotesize{Hyperparameters of the Autoencoder algorithms used for the dataset ECG5000}}
\label{tab:params_ae}
\end{table*}

\begin{table*}[t!]
\begin{center}
\footnotesize
\scalebox{.75}{
\makebox[0.8\textwidth][c]{
\begin{tabular}{c | c c c c c c }
\multicolumn{7}{c}{Variational Autoencoders}\\
\hline
\textbf{Model}  & \textbf{Num Channels} & \textbf{Skip dims} & \textbf{GConv type} & \textbf{Kernel sizes} & \textbf{Dilations} & \textbf{Shrink} \\
\hline
TGraphVAE\_hdcl & $[64]*6,2$ & $[16]*6$ & Sage, $[100]$ & - & - & $35$\\
TGraphMixedVAE\_hdcl* & $[128]*7,8$ & $[32]*7$ & GAT1H & $[3]*7$ & $[1, 2, 4, 8, 16, 32, 64]$ & $35$\\
TCNGraphVAE1\_hdcl & $[128, 112, 96, 80, 64, 48, 8]$ & $[16]*6$ & Sage, $[100]$ & $[16, 12, 8, 4, 2, 2]$ & $[1, 2, 2, 2, 2, 2]$ & $20$\\
TCNGraphVAE2\_hdcl$^\dagger$ & $[128, 112, 96, 80, 64, 48, 8]$ & $[16]*6$ & Sage, $[100]$ & $[16, 12, 8, 4, 2, 2]$ & $[1, 2, 2, 2, 2, 2]$ & $10$\\
\hline
\end{tabular}}}
\end{center}
\caption{\footnotesize{Hyperparameters of the Variational Autoencoder algorithms used for the dataset ECG5000}}
\label{tab:params_vae}
\end{table*}

\section{Results on unsupervised quality detection}\label{app:qdect}
In this section we reproduce the table of the results we obtained in \cite{GSI2025}.

\begin{table*}[t]
\begin{center}
\footnotesize
\scalebox{.84}{
\makebox[1 \textwidth][c]{
\begin{tabular}{c |c c | c c | c | c}
\multirow{2}{*}{\textbf{Model}} & \multicolumn{2}{c|}{Positive class = label $1$} & \multicolumn{2}{c|}{Positive class = label $0$} & \multirow{2}{*}{\textbf{Accuracy}} & \multirow{2}{*}{\textbf{Params}}\\
& \textbf{Precision} & \textbf{Recall} & \textbf{Precision} & \textbf{Recall} &  & \\
\hline
\multicolumn{7}{l}{\emph{Kmeans, approach A}}\\
\hline
TGraphMixedAE & $\textcolor{purple}{0.973} \pm 0.006$ & $0.974 \pm 0.011$ & $0.854 \pm 0.051$ & $\textcolor{purple}{0.847} \pm 0.035$ & $0.955 \pm 0.006$ & $145k$\\
TGraphAE & $0.960\pm 0.007$ & $0.993\pm 0.006$ & $0.952\pm 0.040$ & $0.765\pm 0.044$ & $0.959 \pm 0.003$ & $44k$\\
TCNGraphAE1 & $\textcolor{blue}{0.967}\pm 0.003$ & $\textcolor{blue}{0.998}\pm 0.002$ & $\textcolor{blue}{0.985}\pm 0.014$ & $\textcolor{blue}{0.806}\pm 0.016$  & $\textcolor{purple}{0.969}\pm 0.002$ & $33k$\\
TCNGraphAE2  & $0.965\pm 0.002$ & $0.997 \pm 0.001$ & $0.976\pm 0.007$ & $0.796\pm 0.012$ & $\textcolor{blue}{0.966}\pm 0.002$ & $258k$\\
TCNAE1  & $0.966\pm 0.012$ & $0.995\pm 0.005$ & $0.966 \pm 0.032$ & $0.798\pm 0.074$ & $0.964\pm 0.009$ & $19k$\\
TCNAE2  & $0.949\pm 0.005$ & $\textcolor{purple}{0.999}\pm 0.001$ & $\textcolor{purple}{0.995} \pm 0.006$ & $0.692\pm 0.031$ & $0.954\pm 0.004$ & $208k$\\
\hline
\multicolumn{7}{l}{\emph{Dbscan, approach B}}\\
\hline
TGraphMixedAE & $\textcolor{purple}{0.984} \pm 0.005$ & $0.944 \pm 0.012$ & $0.745 \pm 0.037$ & $\textcolor{purple}{0.909} \pm 0.028$ & $0.939 \pm 0.007$ & $145k$\\
TGraphAE  & $0.968 \pm 0.006$ & $\textcolor{blue}{0.989}\pm 0.006$ & $\textcolor{blue}{0.933}\pm 0.034$ & $0.813\pm 0.038$ & $0.962\pm 0.002$ & $44k$\\
TCNGraphAE1  & $0.971\pm 0.004$ & $\textcolor{purple}{0.991} \pm 0.005$ & $\textcolor{purple}{0.940}\pm 0.028$ & $0.829\pm 0.022$  & $\textcolor{blue}{0.966}\pm 0.003$ & $33k$\\
TCNGraphAE2  & $\textcolor{blue}{0.979}\pm 0.007$ & $0.985 \pm 0.006$ & $0.913\pm 0.031$ & $\textcolor{blue}{0.877}\pm 0.043$  & $\textcolor{purple}{0.967}\pm 0.001$ & $258k$\\
TCNAE1   & $0.971 \pm 0.007$ & $0.985\pm 0.011$ & $0.913 \pm 0.057$ & $0.833 \pm 0.043$ & $0.962\pm 0.007$ & $19k$\\
TCNAE2  & $0.973\pm 0.006$ & $0.988\pm 0.005$ & $0.925 \pm 0.027$ & $0.846\pm 0.039$ & $\textcolor{blue}{0.966}\pm 0.002$ & $208k$\\
\hline
\end{tabular}}}
\end{center}\caption{\footnotesize{Best scores in purple and second best in blue.}}\label{tab2}
\end{table*}
The names of the models are coherent with both the ones introduced in the main body of this paper and the ones found in \cite{GSI2025}. From the table, one can notice that graph-based algorithms slightly outperform the baseline TCN in terms of performance and stability. We refer to our work \cite{GSI2025} for a full description of the experiments and the methodology employed and for the details about approaches A and B that differ mainly as they use Kmeans \cite{KM} or DBscan \cite{DB} to find clusters and assign labels in an unsupervised way in the latent spaces of the autoencoders considered.

\end{document}